\documentclass[10pt,twocolumn,letterpaper]{article}

\usepackage[pagenumbers]{cvpr} 
\usepackage[accsupp]{axessibility}
\usepackage[dvipsnames]{xcolor}

\definecolor{cvprblue}{rgb}{0.21,0.49,0.74}
\usepackage[pagebackref,breaklinks,colorlinks,citecolor=cvprblue]{hyperref}

\definecolor{myBlue}{rgb}{.0, .0, 0.5}
\definecolor{myRed}{rgb}{0.8, .2, .2}
\definecolor{myGreen}{rgb}{0.1, .5, .1}

\def\paperID{2108} 
\def\confName{CVPR}
\def\confYear{2024}

\title{Bilateral Propagation Network for Depth Completion}

\author{
Jie Tang$^1$, \,
Fei-Peng Tian$^2$, \,
Boshi An$^3$, \,
Jian Li$^1$\thanks{indicates the corresponding author.}, \,
Ping Tan$^{4,2}$\\
$^1$National University of Defense Technology, \\
$^2$LightIllusions,  \quad  $^3$Peking University, \\
$^4$The Hong Kong University of Science and Technology\\
}

\begin{document}
\maketitle
\begin{abstract}
    Depth completion aims to derive a dense depth map 
    from sparse depth measurements with a synchronized color image.
    Current state-of-the-art (SOTA) methods are predominantly propagation-based,
    which work as an iterative refinement on the initial estimated dense depth.
    However, the initial depth estimations mostly result from direct applications of 
    convolutional layers on the sparse depth map.
    In this paper, we present a Bilateral Propagation Network (BP-Net), 
    that propagates depth at the earliest stage to avoid directly convolving on sparse data.
    Specifically, our approach propagates the target depth from nearby depth measurements
    via a non-linear model, whose coefficients are generated through a multi-layer perceptron 
    conditioned on both \emph{radiometric difference} and \emph{spatial distance}.
    By integrating bilateral propagation with multi-modal fusion and depth refinement in a multi-scale framework, 
    our BP-Net demonstrates outstanding performance on both indoor and outdoor scenes.
    It achieves SOTA on the NYUv2 dataset 
    and ranks 1st on the KITTI depth completion benchmark at the time of submission.
    Experimental results not only show the effectiveness of bilateral propagation
    but also emphasize the significance of early-stage propagation in contrast to the refinement stage.
    Our code and trained models will be available on the project page.
    \vspace{-1.0 em}
\end{abstract}    
\section{Introduction}

Dense depth perception, \ie estimating per-pixel distance to the camera, is crucial for 3D tasks,
with a rapid growth of applications, such as augmented reality,
 autonomous driving and robotics.
However, due to current hardware limitations,
it is still difficult to directly measure dense depth maps.
Depth measured from LiDAR~\cite{kitti} or SfM~\cite{sfm} is typically sparse,
which is insufficient for real applications like 
scene reconstruction~\cite{rigidfusion} and robot grasping~\cite{robot_grasping}.
Depth completion\footnote{The exact name should be \emph{image guided depth completion}, if considering some early attempts don't utilize color images.},
\ie estimating dense depth map from sparse depth measurements with a synchronized color image,
is a promising solution with the advantage of exploiting
complementary information from data in different modalities.
A key challenge in this task is to effectively process irregularly sampled sparse depth points.

\begin{figure}[t]
   \begin{minipage}{0.225\linewidth}
       \includegraphics[width=1.0\linewidth]{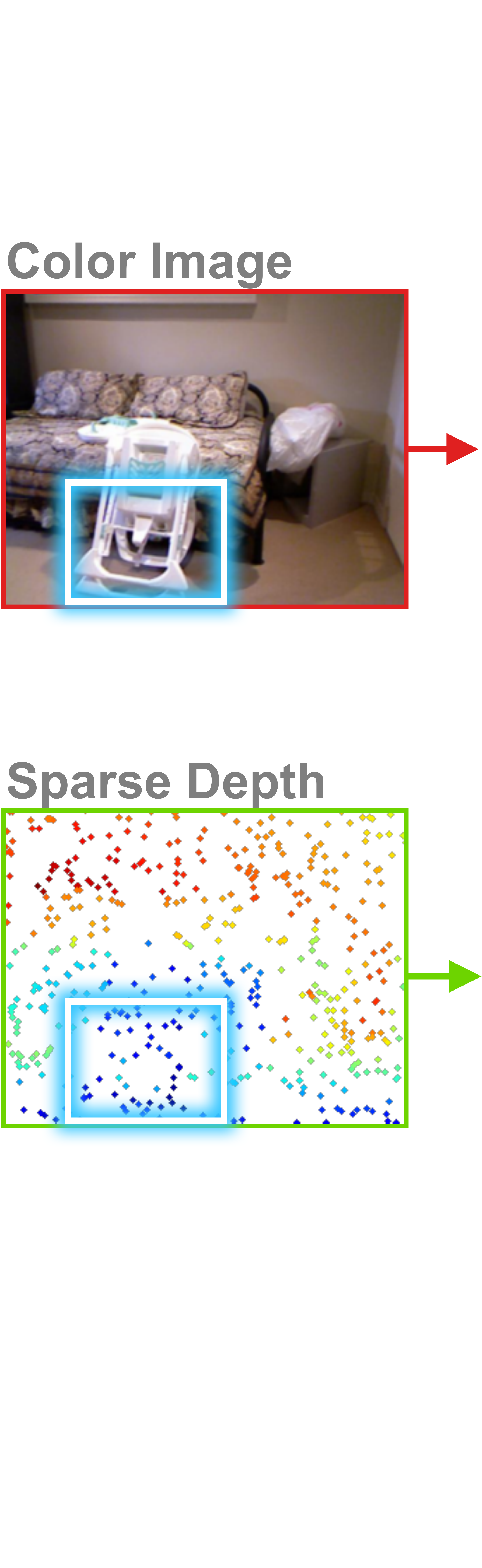}
   \end{minipage}
   \hfill
   \begin{minipage}{0.76\linewidth}
       \subfloat[1-stage depth completion (\eg~\cite{depth_coeff,s2d,guidenet}).]
       {\includegraphics[width=1.0\linewidth]{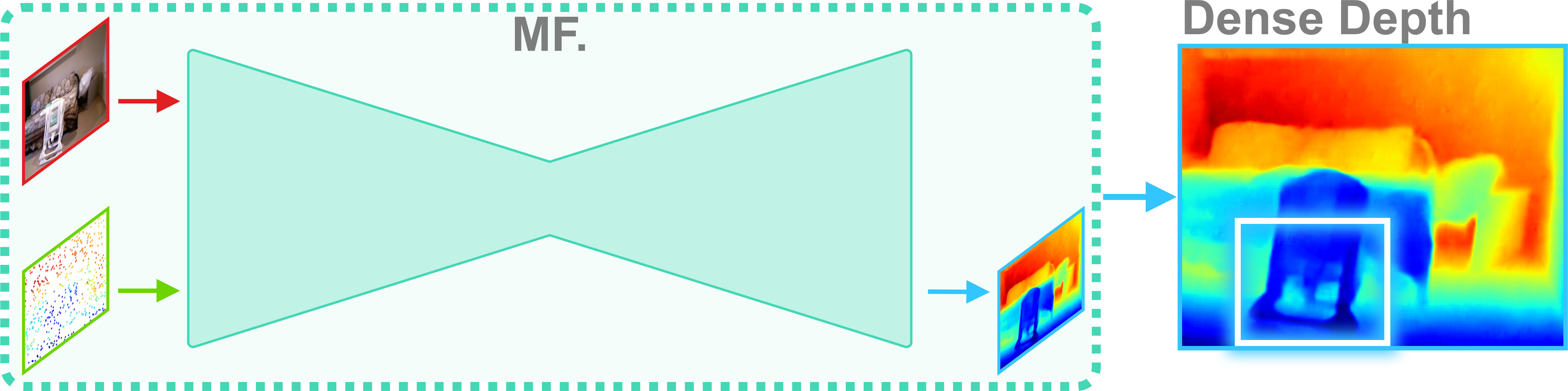}\label{fig:1stages}}
       \vfill
       \subfloat[2-stage depth completion (\eg~\cite{cspn,dyspn,nlspn}).]
       {\includegraphics[width=1.0\linewidth]{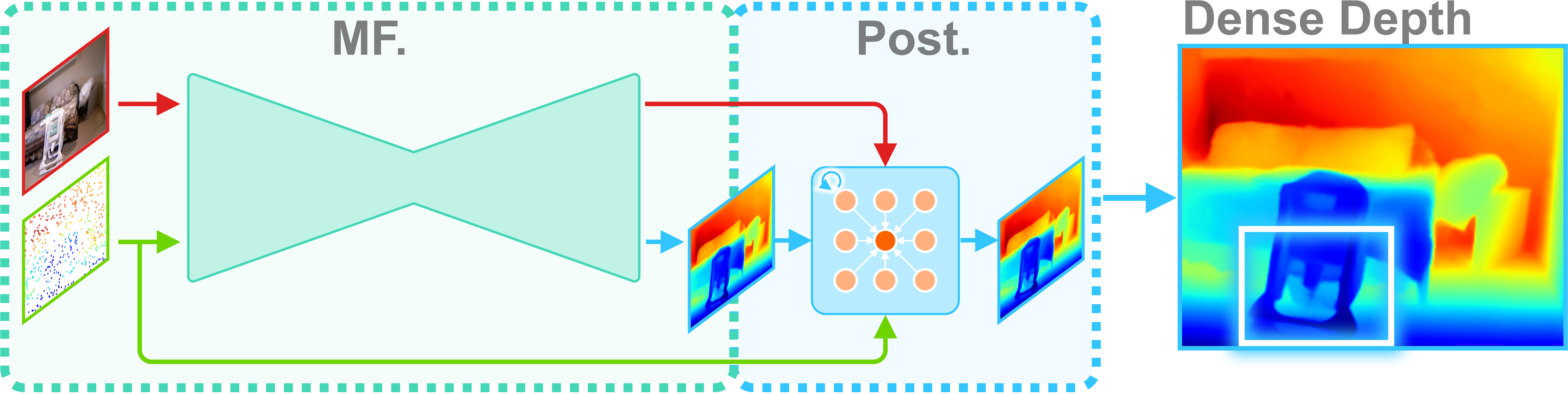}\label{fig:2stages}}
       \vfill
       \subfloat[3-stage depth completion (our BP-Net).]
       {\includegraphics[width=1.0\linewidth]{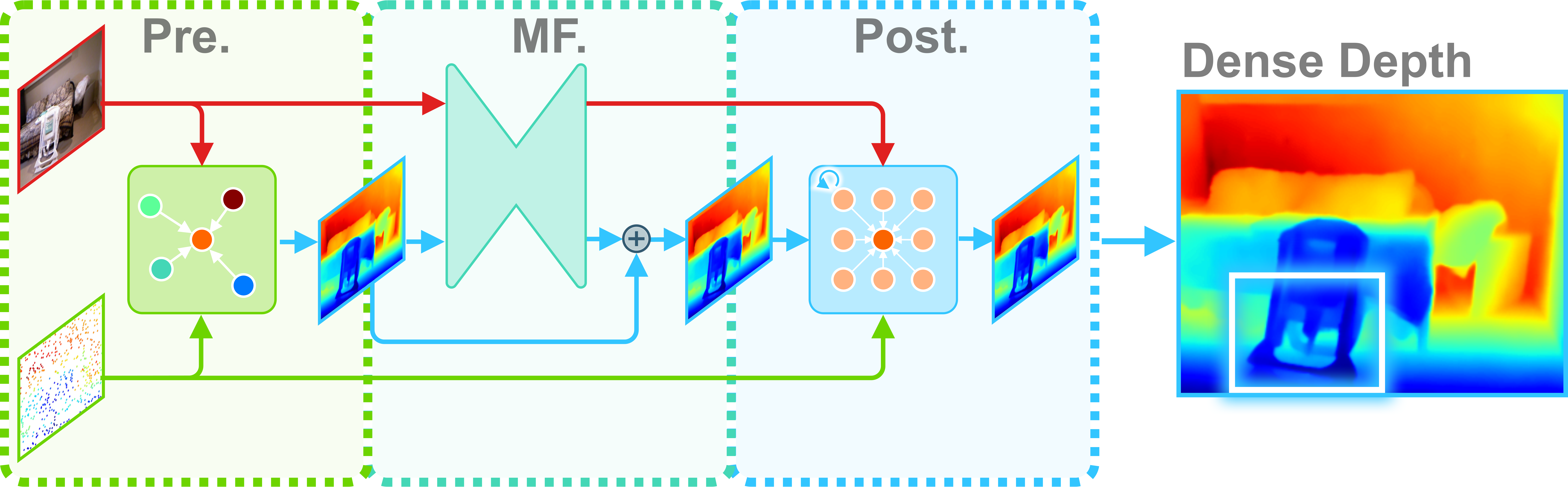}\label{fig:3stages}}
   \end{minipage}
   \caption{Depth completion is to produce a dense depth map from sparse depth measurements with a synchronized color image. (a) Earlier methods take a 1-stage approach to solve this problem by multi-model fusion network (\emph{MF.}). (b) Some recent works include a post-processing stage (\emph{Post.}) to refine the results. (c) Our BP-Net further introduces a pre-processing stage (\emph{Pre.}) to make the multi-modal fusion (\emph{MF.}) network more effective for better results.
   }

\end{figure}

In the context of depth completion,  deep learning-based methods have achieved great success. As shown in~\cref{fig:1stages}, earlier methods~\cite{depth_coeff,s2d,guidenet} often take a 1-stage depth completion strategy by a multi-modal fusion (MF) network to recover dense depth maps. 
These methods mostly suffer from excessive smoothing at depth edges which leads to loss of details in the results.
More recent works~\cite{cspn,dyspn,nlspn} have started to adopt a 2-stage approach as shown in ~\cref{fig:2stages}, with an additional post-processing stage via propagation-based networks.
By revisiting sparse depth measurements to iteratively refine the regressed depth, such propagation-based post-processing can alleviate the over-smoothing problems. Thus, they have better performance and dominate the current state-of-the-art (SOTA).

Although deep learning-based methods generate superior results than conventional methods~\cite{indefense}, these 1-stage and 2-stage methods generally suffer from two common issues. 
Firstly, they typically use $0$ value to indicate an unknown pixel in the input sparse depth map, inducing ambiguity to distinguish between valid and invalid depth measurements as discussed in~\cite{steering_kernel}.
Secondly, the spatially invariant convolution is not 
an ideal processor for irregularly sampled sparse depth points~\cite{SICNN}.
These issues downgrade the performance of the multi-modal fusion network, and the post-processing stage may not be able to solve these issues effectively.

In this paper, we present a bilateral propagation network~(BP-Net) to address the above issues. Our BP-Net is a 3-stage method as illustrated in~\cref{fig:3stages}, with a pre-processing stage to compute an initial dense depth map for the following multi-modal fusion network. The initial dense depth is explicitly propagated from the valid sparse depth measurements, which avoids the above issues of ambiguity and invariant convolution. 
Specifically, we introduce a non-linear propagation model, where the output depth is a combination of nearby valid depth measurements weighted by learned coefficients. Inspired by the well-known bilateral filtering~\cite{bilateral}, we propose a novel multi-layer perceptron to produce these combination coefficients conditioned on both \emph{radiometric difference} and \emph{spatial distance}.
In this way, depth can be propagated with the preference of nearest values in both domain and range. This bilateral propagation is fully differentiable, that can be optimized in an end-to-end fashion together with the multi-modal fusion and depth refinement networks.
Moreover, to deal with long-range depth propagation (\eg a large hole without valid depth measurements nearby), we adopt a multi-scale architecture in which depth is estimated from coarse to fine and the low-resolution result is used as prior for high-resolution estimation. Furthermore, a multi-scale loss is designed to better supervise the multi-scale network.

We perform experiments to validate our BP-Net with both indoor and outdoor scenes, under standard evaluation criteria. Our method achieves SOTA performance on the NYUv2 dataset and ranks 1st on the KITTI benchmark at the time of paper submission.
We conduct ablation studies to demonstrate the efficacy of each component and also analyze performance under various sparsity levels. Experimental results not only show the effectiveness of bilateral propagation but also emphasize the significance of early-stage propagation in contrast to the refinement stage. Code and trained models are available at \url{https://github.com/kakaxi314/BP-Net}.


\section{Related Work}

To estimate dense depth map, 
many techniques work on data from RGB-D camera, like Kinect,
whose goal is depth super-resolution or depth inpainting.
To this end,
filter-based~\cite{bilateral_upsampling,guide_filter,mean_filter}, 
optimization-based~\cite{markov_field,autoregression,TV}, 
and learning-based~\cite{sparse_repr,joint_filter,DCRGBD} methods are proposed to usample resolution
or fill holes in depth map.
On the other hand, depth completion 
from irregularly distributed extremely sparse data measured by LiDAR or SfM 
attracts increasing attention with a large number of approaches emerging in recent years.
In general,
previous depth completion approaches mainly focused on the following three problems:
how to process sparse data, how to fuse multi-modal data, 
and how to refine results.
We briefly review previous techniques from these three aspects.

\textbf{Sparse Data Processing.} 
As traditional methods,
Ku~\etal~\cite{indefense} utilize a series of classical image processing operators
like dilation, hole filling \etc to densify sparse depth map.
Zhao~\etal~\cite{surface_geometry} calculate surface normal in spherical coordinate system,
and then estimate depth with local surface sharing assumption.
In deep learning era, 
some early attempts focus on sparsity invariant operations to avoid simply convolving on sparse depth map.
Uhrig~\etal~\cite{SICNN} replace the conventional convolution operation 
with sparsity-invariant convolution that keeps track of validation masks at each layer.
Huang~\etal~\cite{HMSnet} employ the sparsity-invariant layer in an encoder-decoder network structure.
Eldesokey~\etal~\cite{confprop} further extend the validation masks in sparsity invariant layer to 
 a continuous confidence field.
There are also works embedding sparse depth by differentiable solver in an end-to-end trained network.
Qu~\etal~\cite{basis_fitting, bb_fitting} process sparse depth by solving linear regression in deep network.
Conti~\etal~\cite{SADC} optimize weighted linear regression in multi-scale to inject sparse depth into network.
Some works propose to first densify the sparse
depth by classical approaches, and then enhance the densified depth by learning-based method.
Chen~\etal~\cite{nearest_interp} adopt nearest interpolation on sparse depth map
 before feed it to deep network.
Wang~\etal~\cite{lrru} utilize~\cite{indefense} to estimate the initial depth map.
Liu~\etal~\cite{steering_kernel} further replace the hand-crafted interpolation with
a differentiable kernel regression layer.
In comparison, our BP-Net 
propagates an initial dense depth map for the following operations, 
but depending on both image content and spatial distance, 
while spatial distance from valid depth to the target pixel is not considered in previous methods.

\textbf{Multi-modal Fusion.}  
To fuse multi-modal data of depth map and color image,
some methods~\cite{s2d,ss2d,depth_coeff} adopt the early-fusion scheme, 
\eg Ma \etal~\cite{s2d} concatenate the depth map and the RGB image as input for deep network to process.
While some~\cite{guidenet,ddp,rignet} adopt the late-fusion scheme,
\eg Tang~\etal~\cite{guidenet} propose a guided convolutional module to fuse image feature with depth feature in multi-stage.
Some works~\cite{deeplidar,normal_constraint} fuse depth map with estimated surface normal, 
\eg Qiu~\etal~\cite{deeplidar}  utilize surface normal as intermediate representation for fusion with sparse depth.
And some~\cite{2d3d,acmnet} explore the geometry information in 3D points, 
\eg Chen~\etal~\cite{2d3d} fuse 2D image feature with 3D point feature for depth estimation.
Recent, Transformer architecture~\cite{attention} 
is also introduced to establish attention mechanism for multi-modal fusion~\cite{guideformer,cformer}.
Our BP-Net mainly focused on sparse data processing stage, 
with only a simple U-Net structure for multi-modal fusion in early-fusion scheme, 
and is not conflict to other multi-modal fusion designs in theory.

\textbf{Depth Refinement.} 
Directly regressed depth map may suffer from blur effect on object boundaries.
Depth refinement mostly follows the spatial propagation mechanism~\cite{spn},
which iteratively refines the regressed depth by a local linear model with learned affinity.
Cheng~\etal~\cite{cspn} adopt convolution operation to update all pixels simultaneously with a fixed kernel size, 
and further extend it to CSPN++~\cite{cspn++} 
assembling results from multiple kernel sizes and multiple iteration steps.
Park~\etal~\cite{nlspn} provide a non-local neighborhood for propagation
by learning the offsets to the regular grid.
Lin~\etal~\cite{dyspn} extend the propagation process 
with affinity matrix adaptive to neighbors of different distances.
Liu~\etal~\cite{graphcspn} construct graph neural network integrating 3D information 
to estimate neighbors in each update iteration.
Wang~\etal~\cite{lrru} adjust the kernel scope from large to small during the
update process.
These propagation-based methods dominate current SOTA,
but still require an initial dense depth which is mostly yielded by
directly applying convolutional layers on the sparse depth map.
In contrast, our BP-Net propagates depth at the earliest stage to avoid convolving on sparse depth map.
Though we implement a simple CSPN++~\cite{cspn++} for depth refinement,
our method should be also compatible to other depth refinement approaches.

\section{The Proposed Method}

\subsection{Overview}

\begin{figure*}[t]
    \begin{center} 
       \includegraphics[width=0.98\textwidth]{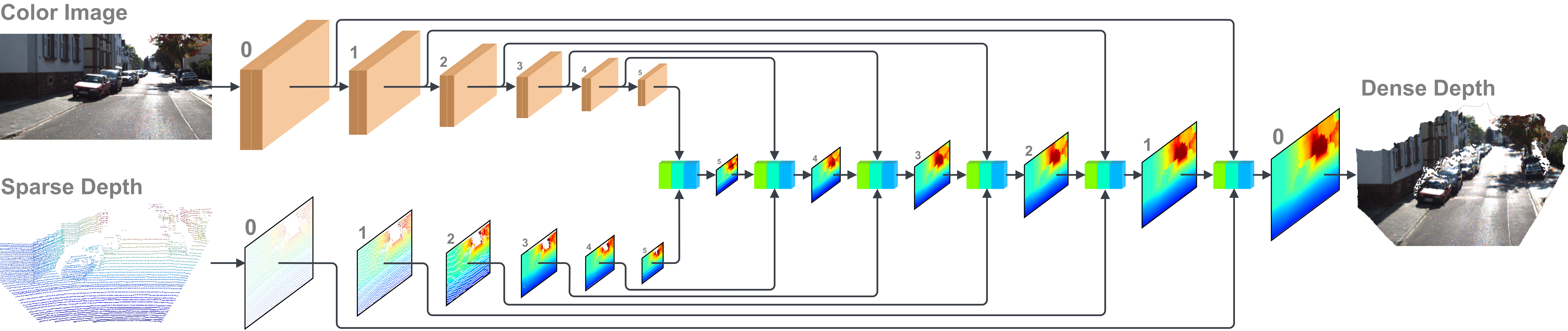}
    \end{center}
       \caption{\textbf{Overview of the proposed approach.}
       Dense depth is estimated in a multi-scale scheme, from low-resolution to high-resolution.
       In each scale, dense depth map is produced from sparse depth map and image feature by 3-stage processing depicted in~\cref{fig:3stages}.
       }\label{fig:overview}
       \vspace{-1em}
\end{figure*}

Let $I$ denote a color image, and $S$ represent the synchronized sparse depth map typically obtained by projecting measured 3D points onto the image plane using calibration parameters. The color image $I$ provides rich scene context with well-defined boundaries and precise semantics. In contrast, the depth map $S$, while sharing the same resolution as $I$, consists of sparse and irregular yet fairly accurate depth measurements with definite geometry. The goal of depth completion is to produce a dense depth map $D$ by exploring the complementary information embedded in $I$ and $S$.

As shown in~\cref{fig:overview},
the proposed method is a multi-scale network with $6$ scales marked from $0$ to $5$. The dense depth $D$ is estimated from the lowest resolution in scale $5$ to the highest resolution in scale $0$. 
For any scale $s$ in this network, the dense depth map is produced in three sequential stages as illustrated in~\cref{fig:3stages}:  a pre-processing stage to generate an initial dense depth map from sparse depth measurements via the proposed bilateral propagation module (in~\cref{sec:bp_module}), a multi-modal fusion stage to generate a residual depth map by a simple U-Net fusing both image features and depth features in an early fusion scheme (in~\cref{sec:multi_fusion}), and a refinement stage to update the dense depth map with sparse depth via convolutional propagation module in an iterative manner (in~\cref{sec:depth_refine}).
Finally, we also introduce the implementation details in~\cref{sec:imp_detail}.

\begin{figure}[t]
    \begin{center} 
       \includegraphics[width=0.98\columnwidth]{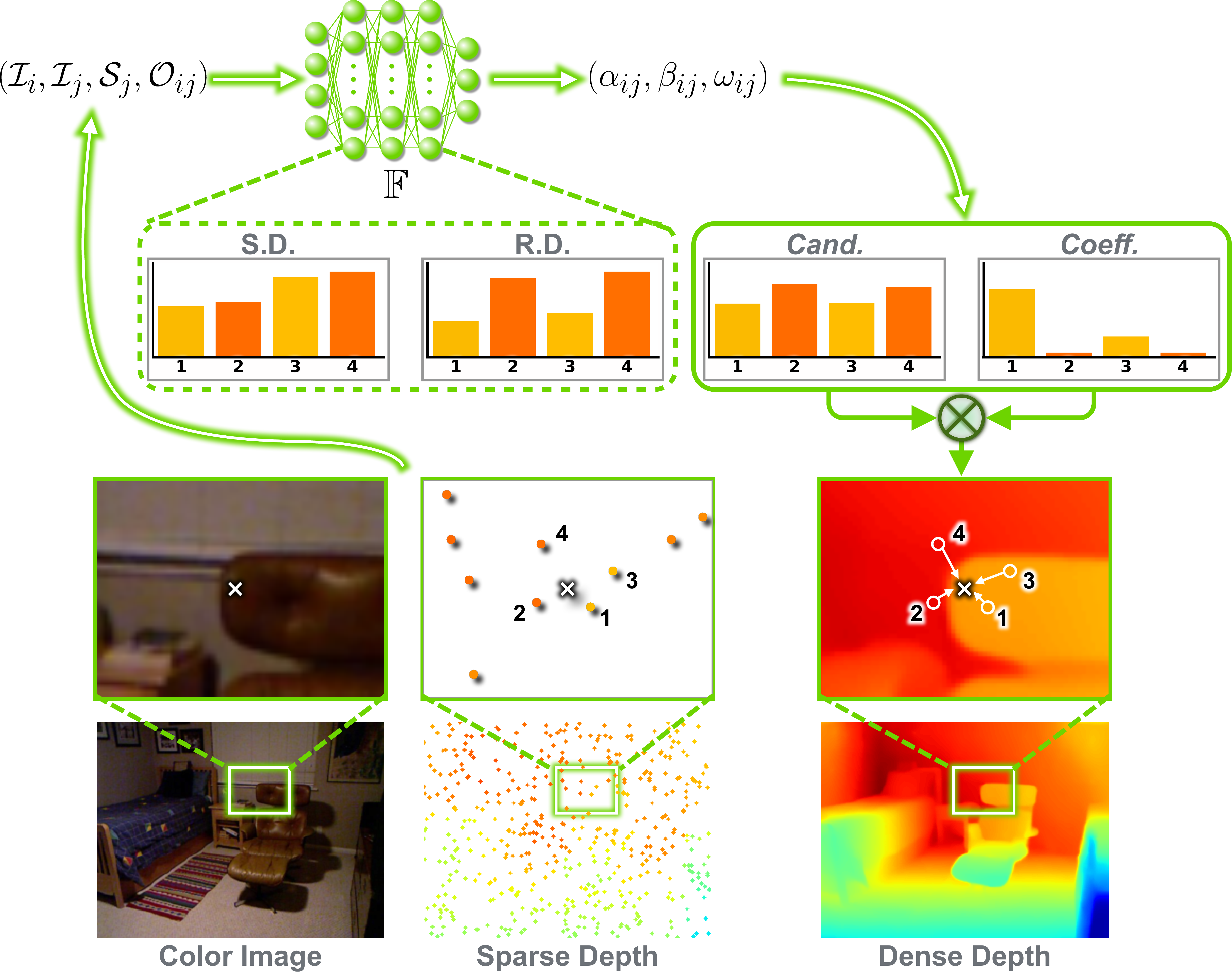}
    \end{center}
       \caption{\textbf{Bilateral Propagation Module.}
       The target depth is a convex combination of 
        depth candidates (\emph{Cand.}) weighted by coefficients (\emph{Coeff.}),
        which are yielded by a MLP $\mathbb{F}$ conditioned on both 
        spatial distance (\emph{S.D.}) and radiometric difference (\emph{R.D.})
        from the nearby depth measurements.}\label{fig:module}
\end{figure}

\subsection{Bilateral Propagation Module} \label{sec:bp_module}

\subsubsection{Depth Parameterization}

We aim to propagate a dense depth map ${D}'$ from the sparse depth map $S$ to mitigate the issues arising from the sparsity problem in the subsequent multi-modal fusion.
Instead of na\"{\i}vely applying convolution operations on the irregular sparse depth map,
 we explicitly model the depth propagation process as:
\begin{equation}\label{eq:dep_para}
    {{D}'}_i = \sum_{j \in \mathcal{N}(i)} \omega_{ij} {{D}'}_{ij}^{c} = \sum_{j \in \mathcal{N}(i)} \omega_{ij}(\alpha_{ij} S_j+\beta_{ij}).
\end{equation} 
Here, the target depth ${{D}'}_i$ at pixel $i$ is a local combination of $N$ nearest valid sparse depth $S_j$ with coefficients $\omega_{ij}$, $\alpha_{ij}$ and $\beta_{ij}$, 
where $ j \in \mathcal{N}(i) $ is a nearby pixel. 
Specifically, our approach first generate depth candidates ${{D}'}_{ij}^{c}$ 
by an affine transformation of the sparse depth $S_j$ with coefficients $\alpha_{ij}$ and $\beta_{ij}$, 
then linearly combine depth candidates ${{D}'}_{ij}^{c}$ with coefficient $\omega_{ij}$ to produce the target depth ${{D}'}_i$. We adopt Euclidean distance on the image plane to find $\mathcal{N}(i)$ containing $N$ nearest neighboring valid pixels for any pixel $i$. 
Notably, we observe that a relatively small number is sufficient for $N$. Based on empirical findings, we set $N=4$. Further discussion about the choice of $N$ can be found in~\cref{sec:local_num}.

Our depth parameterization is a general formulation for dense depth estimation from sparse valid depth. Many conventional interpolation methods can be viewed as specific cases of our formulation. For instance, nearest interpolation corresponds to the case with $N=1$, $\omega_{ij}=1$, and bilateral interpolation aligns with a specific case where $\omega_{ij}$ is determined by a handcrafted kernel. 
Moreover, in our formulation, $\alpha$, $\beta$, and $\omega$ are dynamically generated depending on both image content and spatial distance, providing adaptability to different sparse and irregular depth distributions. For cases where the depth at the target pixel is far away from those at the neighboring valid pixels, our formulation might learn an $\alpha$ close to 0, disregarding the neighboring valid depth and directly regressing the target depth through the parameter $\beta$. This scenario is challenging for manually designed local filters to address.
Consequently, our depth parameterization is much more powerful than a mere local filter on sparse valid depth and thus can help avoid the sparsity issue in the following network operations.

\subsubsection{Parameter Generation}

To obtain the above coefficient parameters $\alpha$, $\beta$, and $\omega$ for depth propagation,
we draw inspiration from the well-established bilateral filtering~\cite{bilateral}. Bilateral filtering~\cite{bilateral} is known for producing edge-preserving results by combining both \textbf{radiometric difference} and \textbf{spatial distance} in generating filtering weights. As illustrated in \cref{fig:module}, we adopt a MLP as the hyper-network $\mathbb{F}$, that is
\begin{equation} \label{eq:para_gen}
    \alpha_{ij},  \beta_{ij}, \omega_{ij} = \mathbb{F}(\mathcal{I}_{i},\mathcal{I}_{j},\mathcal{S}_{j},\mathcal{O}_{ij}).
\end{equation}
The $\mathbb{F}$ is conditioned on prior encodings including the image encoding $\mathcal{I}$ at target pixel $i$ and source pixel $j$, the depth encoding $\mathcal{S}$ at source pixel $j$, and the spatial offset encoding $\mathcal{O}$ for pixel coordinates from $i$ to $j$.
The spatial offset encoding $\mathcal{O}$ serves as the \textbf{spatial distance} term. Instead of designing an explicit radiometric difference term, we use MLP $\mathbb{F}$, to \textbf{implicitly} consider the \textbf{radiometric difference} with image encoding $\mathcal{I}_{i}$ and $\mathcal{I}_{j}$ as input. 
We believe our formulation can further explore additional information like scene context, which goes beyond
the traditional bilateral filter that utilizes an explicitly designed radiometric difference term, although our initial inspiration
comes from it.

The adopted MLP consists of four densely connected layers, 
each followed by a Batch Normalization~\cite{batchnorm} and a GeLU~\cite{gelu} layer,
with a skip connecting layer adding the outputs from the second and the last layer.
In practice, $\alpha$, $\beta$ are directly regressed.
For any target pixel~$i$, $\omega$ is obtained from an extra added Softmax layer. The layer operates among pixel~$i$'s $N$ neighboring valid pixels to ensure that $\sum_{j \in \mathcal{N}(i)} \omega_{ij}=1$.
This MLP is shared among any $(i,j)$ pixel pair but takes different encodings as input to generate \emph{spatial-variant} and \emph{content-dependent} parameters.

The paradigm of \emph{content-dependent} parameter generation has been used in previous depth completion methods and gained robust results. GuideNet~\cite{guidenet} and RigNet~\cite{rignet}
leverage \emph{content-dependent} weights to establish a sophisticated multi-modal feature fusion.
CSPN~\cite{cspn} and CSPN++~\cite{cspn++} dynamically generate \emph{content-dependent} affinity matrix to iteratively enhance the initial depth.
There are two notable differences between these existing methods and ours.
Firstly, the \emph{content-dependent} parameters are only learned from image content in these methods, typically neglecting the spatial distance between the target pixel and nearby depth measurements. In contrast, 
our method takes both image content and spatial distance into account. This allows the propagation of depth with the preference for nearest values in both content and spatial domains.
Secondly, the \emph{content-dependent} parameters are typically employed in the multi-modal fusion or refinement stage in existing methods while we extend this paradigm to the earliest pre-processing stage, enabling subsequent stages to overcome the sparsity problem more effectively.

\subsubsection{Prior Encoding}\label{sec:prior_encoding}

The proposed bilateral propagation module is arranged in a multi-scale scheme, 
with prior encodings from the corresponding resolution.
For the lowest resolution with $s=5$,
we directly adopt the image feature $\mathbf{I}^{s}$ as image encoding.
Otherwise, to make image encoding representative, 
we concatenate the multi-modal fused feature $\mathbf{F}^{s+1}$ 
(introduced in \cref{sec:multi_fusion}) with depth feature $\mathbf{D}^{s+1}$ in scale $s+1$.
The depth feature $\mathbf{D}^{s+1}$ is achieved by inverse projecting estimated depth $D^{s+1}$ to camera space.
Then we utilize deconvolution operation to upsample the concatenated feature map to scale $s$.
Finally, we concatenate the upsampled feature with $\mathbf{I}^{s}$, 
and adopt an extra convolution operation to produce the image encoding $\mathcal{I}^{s}$.

We inverse project sparse depth map $S^{s}$ to camera space as depth encoding $\mathcal{S}^{s}$. 
Except $S^{0}=S$, 
the sparse depth map $S^{s}$ is downsampled from $S$ via a weighted pooling, whose weights are generated from the image encoding $\mathcal{I}^s$.
Specially,
we adopt a periodic shuffling operator to rearrange the image feature map 
from shape ${\frac{H}{2^s}\times \frac{W}{2^s} \times 4^{s}}$ to shape $H\times W$
and apply the exponential transformation,
that guarantees the generated weight map is positive and has the same resolution as $S$. 
More implementation details are introduced in the supplementary material.
While the points in low-resolution depth maps are denser, we only use the valid ones for depth propagation shown in \cref{eq:dep_para} and \cref{eq:para_gen}.

We adopt the directed distance from the target pixel to the valid source pixel,
\ie a 2D vector representing offsets along two axes of the image plane,
 as the spatial distance encoding $\mathcal{O}$.
Position encoding, which transforms the position to a high dimensional space via Fourier transform or learned embedding, has been used in some recent popular network structures, \eg Transformer~\cite{attention} and NeRF~\cite{nerf}. We find high dimensional position encoding can induce little performance improvement
in our formulation, despite the increased computational cost.
We attribute this to the simplicity of a spatial distance term being adequate for coefficient generation in our local combination model, akin to bilateral filtering.

\subsection{Multi-Modal Fusion} \label{sec:multi_fusion}
For each scale, we can get a dense depth map ${{D}'}^{s}$ from the bilateral propagation module
 which is suitable for CNN to process.
We adopt image encoding $\mathcal{I}^{s}$ as the image feature,
and inverse project ${{D}'}^{s}$ to camera space as the depth feature.
Then, we follow the early-fusion scheme that simply concatenates the image feature and depth feature,
and feeds the concatenated feature as input to a U-Net structure~\cite{unet} for multi-modal fusion.
The adopted U-Net is an encoder-decoder CNN, aggregating local and global features in multi-scale.
The encoder part consists of 2 ResNet blocks~\cite{resnet} for feature extraction in each scale and 
convolution layer with stride 2 to reduce the resolution of the feature map.
The decoder part consists of a deconvolution layer with stride 2 to upsample the feature map
and skip connection to fuse the upsampled feature and encoded feature with the same resolution by concatenation operation.
The U-Net finally yields a fused feature map $\mathbf{F}^{s}$.
And a convolution layer is applied on $\mathbf{F}^{s}$ to estimate a residual depth map, which is then added to the previous depth map ${{D}'}^{s}$, 
resulting in a processed depth map ${{D}''}^{s}$.

\subsection{Depth Refinement} \label{sec:depth_refine}

We implement a simple convolutional spatial propagation module similar to CSPN++~\cite{cspn++} as the depth refinement for each scale.
This propagation process updates the processed depth map ${D}''$ aided by sparse valid depth map $S$.
The updating equation at step $t$ for a preset propagation kernel size $k$ can be written as:

\begin{equation} \label{eq:conv_prop}
    \begin{split}
        \begin{matrix}
            &\hat{D}_{i,k,t}=\kappa_{i,k} \hat{D}_{i,k,t-1} + \sum\limits_{j \in \mathcal{N}_{k}(i) \setminus i} \kappa_{j,k} \hat{D}_{j,k,t-1}, \\
            &\kappa_{i,k}= 1-\sum\limits_{j \in \mathcal{N}_{k}(i) \setminus i} \kappa_{j,k}, \\
            &\kappa_{j,k}= \frac{\hat{\kappa}_{j,k}}{\sum\limits_{j \in \mathcal{N}_{k}(i) \setminus i} \left | \hat{\kappa}_{j,k}\right |},
        \end{matrix}
    \end{split}
\end{equation}
where $\kappa$ is the content-dependent affinity map 
and $\hat{\kappa}$ is generated by convolutional layers depending on the fused feature $\mathbf{F}$.
$\hat{D}_{t}$ is initialized with ${D}''$ for $t=0$.
The $l^{1}$-norm constraint for $\hat{\kappa}$ guarantees the stability of propagation process~\cite{cspn}.
Note that $\mathcal{N}_{k}$ used here is different from 
$\mathcal{N}$ used in our bilateral propagation module.
$\mathcal{N}_{k}$ is the set of neighboring pixels in a $k\times k$ local window, irrelevant to the validation of sparse depth measurement.

For each refinement step, we also update the propagated depth $\hat{D}_{t}$ by embedding sparse depth measurement
 after performing \cref{eq:conv_prop},
\begin{equation}
    \hat{D}_{i,k,t} \leftarrow (1-\gamma_{i,k}\mathbb{I}(S_{i}))\hat{D}_{i,k,t} + \gamma_{i,k}\mathbb{I}(S_{i})S_{i},
\end{equation}
where $\gamma$ works as confidence generated by convolutional layers followed by a sigmoid layer
based on the fused feature $\mathbf{F}$.
The whole process iterates $T$ steps, and
the final depth is a combination of multi-kernel and multi-step:
\begin{equation}
    D_{i} = \sum_{t \in \mathcal{T}} \sum_{k \in \mathcal{K}}  \tau_{i,t}\sigma_{i,k}\hat{D}_{i,k,t}.
\end{equation}
Here, $\mathcal{K}$ is a set of $\{3,5,7\}$ for 3 different kernel sizes, 
$\mathcal{T}$ is a set of $\{0,\lfloor T/2 \rfloor,T\}$ indicating different iteration steps.
$\tau$ and $\sigma$ work as confidence maps generated by convolutional layers followed by a Softmax layer, 
normalized across different iteration steps and different kernel sizes respectively.
Empirically, we set a low iteration number $T$ for refinement on low-resolution depth map, 
that $T$ is from $2$ to $12$ with an incremental step of $2$ for a total of $6$ scales from low-resolution to high-resolution.

\subsection{Implementation Details} \label{sec:imp_detail}

\subsubsection{Loss Function}
Our method is trained in an end-to-end manner, with a multi-scale loss to 
provide adequate supervision on the depth map estimated in each scale.
The loss function is
\begin{equation} \label{eq:loss}
    L = \sum_{s=0}^{5} \sum_{i \in \mathcal{P}_{v}} \lambda_{s}  \| D_{i}^{gt}-{\mathbb{U}_{s}(D^{s})}_{i}\|^{2},
    \end{equation}
where $\mathcal{P}_{v}$ represents the set of valid pixels in the ground truth depth map $D^{gt}$.
$\mathbb{U}_{s}$ is bilinear interpolation operation to upsample predicted depth map in scale $s$ 
to the same resolution as $D^{gt}$,
and $\lambda_{s}$ is hyper-parameter to balance loss in each scale, which is set as $4^{-s}$ empirically.

\subsubsection{Training Setting}

We implement our method in pytorch~\cite{pytorch} and train it on a GPU workstation with 4 Nvidia RTX 3090 GPUs.
Our network is mainly stacked with ResNet blocks~\cite{resnet}, in which a DropPath~\cite{droppath} is added before residual addition as a regularization for training.
We adopt AdamW~\cite{adamW} with 0.05 weight decay as the optimizer, 
and clip gradient whose $l^2$-norm is larger than $0.1$.
Our method is trained from scratch in roughly $300K$ iterations 
with OneCycle learning rate policy~\cite{onecycle} gradually 
reducing the learning rate to $25\%$ of the largest learning rate.
We set batch size as $8$ and the largest learning rate as $0.001$ for KITTI dataset.
For NYUv2 dataset, whose image resolution is lower,
we increase the batch size to $16$ and the largest learning rate to $0.002$ correspondingly.
The final model is yielded by Exponential Moving Average (EMA) with $0.9999$ decay.

\section{Experiments}

\subsection{Datasets \& Evaluation Metrics}\label{sec:dataset}

We conduct comprehensive experiments to verify our method on both indoor and outdoor scenes.

\textbf{Indoor scene}. The NYUv2 dataset~\cite{nyuv2} is an
indoor dataset containing 464 scenes gathered by a Kinect sensor. We take $50K$ frames sampled from 249 scenes by Ma~\etal~\cite{s2d} as the training data, and evaluate on the official test set consisting of 654 samples from 215 scenes.
Following the common practice~\cite{guidenet,nlspn,cformer}, images are down-sampled to $320\times240$ and then center-cropped to $304\times228$, and for each frame, sparse depth is generated by random sampling $500$ points from the ground truth depth map.
Due to the input resolution for our network being a multiple of $32$, we further pad the image to $320\times256$ as input, but evaluate only the valid region of size $304\times228$ to keep a fair comparison with other methods. The standard evaluation metrics are root mean squared error (RMSE), mean absolute relative error (REL), and $\delta_{\theta}$ represents the percentage of pixels whose error is less than a threshold $\theta$.

\begin{table}[t]
    \begin{center}
    \scalebox{0.60}{
    \begin{tabular}{@{\extracolsep{3pt}}lccccccc@{}} \toprule
        &  \multicolumn{4}{c}{KITTI} &  \multicolumn{3}{c}{NYUv2}  \\ \cline{2-5} \cline{6-8}
     & \begin{tabular}[c]{@{}c@{}}\textbf{RMSE}$\downarrow$ \\ ($mm$)\end{tabular}  & \begin{tabular}[c]{@{}c@{}}MAE$\downarrow$ \\ ($mm$)\end{tabular}  & \begin{tabular}[c]{@{}c@{}}iRMSE$\downarrow$ \\ ($1/km$)\end{tabular} & \begin{tabular}[c]{@{}c@{}}iMAE$\downarrow$ \\ ($1/km$)\end{tabular} & \begin{tabular}[c]{@{}c@{}}\textbf{RMSE}$\downarrow$ \\ ($m$)\end{tabular}  & \begin{tabular}[c]{@{}c@{}}REL$\downarrow$ \\ $\ $ \end{tabular} & \begin{tabular}[c]{@{}c@{}}$\delta_{1.25}\uparrow$ \\ ($\%$)\end{tabular} \\ \midrule
    S2D~\cite{s2d} & 814.73 & 249.95 & 2.80 & 1.21 & 0.230 & 0.044 & 97.1 \\
    CSPN~\cite{cspn} &  1019.64 & 279.46 & 2.93 & 1.15 & 0.117 & 0.016 & 99.2 \\
    DeepLiDAR~\cite{deeplidar} & 758.38 & 226.50 & 2.56 & 1.15 & 0.115 & 0.022 & 99.3 \\
    CSPN++~\cite{cspn++}  & 743.69 & 209.28 & 2.07 & 0.90 & 0.115 & -- & -- \\
    GuideNet~\cite{guidenet} & 736.24 & 218.83 & 2.25 & 0.99 & 0.101 &0.015&99.5\\
    FCFR~\cite{penet} & 735.81&  217.15&  2.20&  0.98  & 0.106 & 0.015& 99.5 \\
    ACMNet~\cite{acmnet} & 744.91 & 206.09 & 2.08 &  0.90 & 0.105 & 0.015 & 99.4 \\
    NLSPN~\cite{nlspn} & 741.68 & 199.59 & 1.99 & 0.84 &  0.092 & 0.012 & \textbf{99.6} \\
    RigNet~\cite{rignet} & 712.66 & 203.25 & 2.08 & 0.90 & 0.090 & 0.013 & \textbf{99.6} \\
    DySPN~\cite{dyspn} & 709.12 & 192.71 & 1.88 & 0.82 & 0.090 &  0.012 & \textbf{99.6}\\
    BEV@DC~\cite{bevdc} & 697.44 & \textbf{189.44} & 1.83 & 0.82 & \textbf{0.089} &0.012& \textbf{99.6} \\
    CFormer~\cite{cformer} &  708.87 & 203.45 & 2.01 & 0.88 & 0.090 &0.012&-- \\
    LRRU~\cite{lrru} & 696.51 & 189.96 & 1.87 & \textbf{0.81} &  0.091 & \textbf{0.011} & \textbf{99.6} \\
    \midrule
    BP-Net & \textbf{684.90} & 194.69 & \textbf{1.82} & 0.84  & \textbf{0.089} & 0.012 & \textbf{99.6} \\ \bottomrule
    \end{tabular}
    }
    \vspace{-0.5 em}
    \caption{{\bf Performance on KITTI and NYUv2 datasets.}
    For the KITTI dataset, results are evaluated by the KITTI testing server and ranked by the RMSE (in $mm$).
    For the NYUv2 dataset, authors report their performance on the official test set in their papers.
    The best result under each criterion is in \textbf{bold}.
    } \label{tab:comparison}
    \end{center}
    \vspace{-2em}
 \end{table}

\textbf{Outdoor scene}. The KITTI depth completion (DC) dataset~\cite{SICNN} is collected by an autonomous driving vehicle, whose ground truth depth is from temporally registered LiDAR scans, further verified by stereo image pairs. 
This dataset provides $86,898$ and $1,000$ frames for training and validation respectively. Another $1,000$ frames for testing are evaluated on the remote server with a public leaderboard\footnote{\url{http://www.cvlibs.net/datasets/kitti/eval_depth.php?benchmark}} for ranking. We randomly crop frames to $1216 \times 256$ for training and directly use the full-resolution frames as input for testing. The standard evaluation metrics are root mean squared error (RMSE), mean absolute error (MAE), root mean squared error of the inverse depth (iRMSE), and mean absolute error of the inverse depth (iMAE),
among which RMSE is chosen as the primary metric for ranking.

\subsection{Comparison with SOTAs}

We evaluate our method on the test set of the NYUv2 dataset and KITTI DC dataset separately. 
\cref{tab:comparison} lists the quantitative comparison of our method and other top-ranking published methods. On the KITTI DC leaderboard, our method ranks 1st and exceeds all other methods under the primary RMSE metric at the time of paper submission.
It also has comparable performance under other evaluation metrics. On the NYUv2 dataset, our method achieves the best RMSE and the best $\delta_{1.25}$. Due to the diverse viewpoints and sparser input depth map in the NYUv2 dataset, it is more challenging to obtain a large performance gain in the NYUv2 dataset compared with the KITTI dataset.
\begin{figure*}
    \centering
    \includegraphics[width=0.97\textwidth]{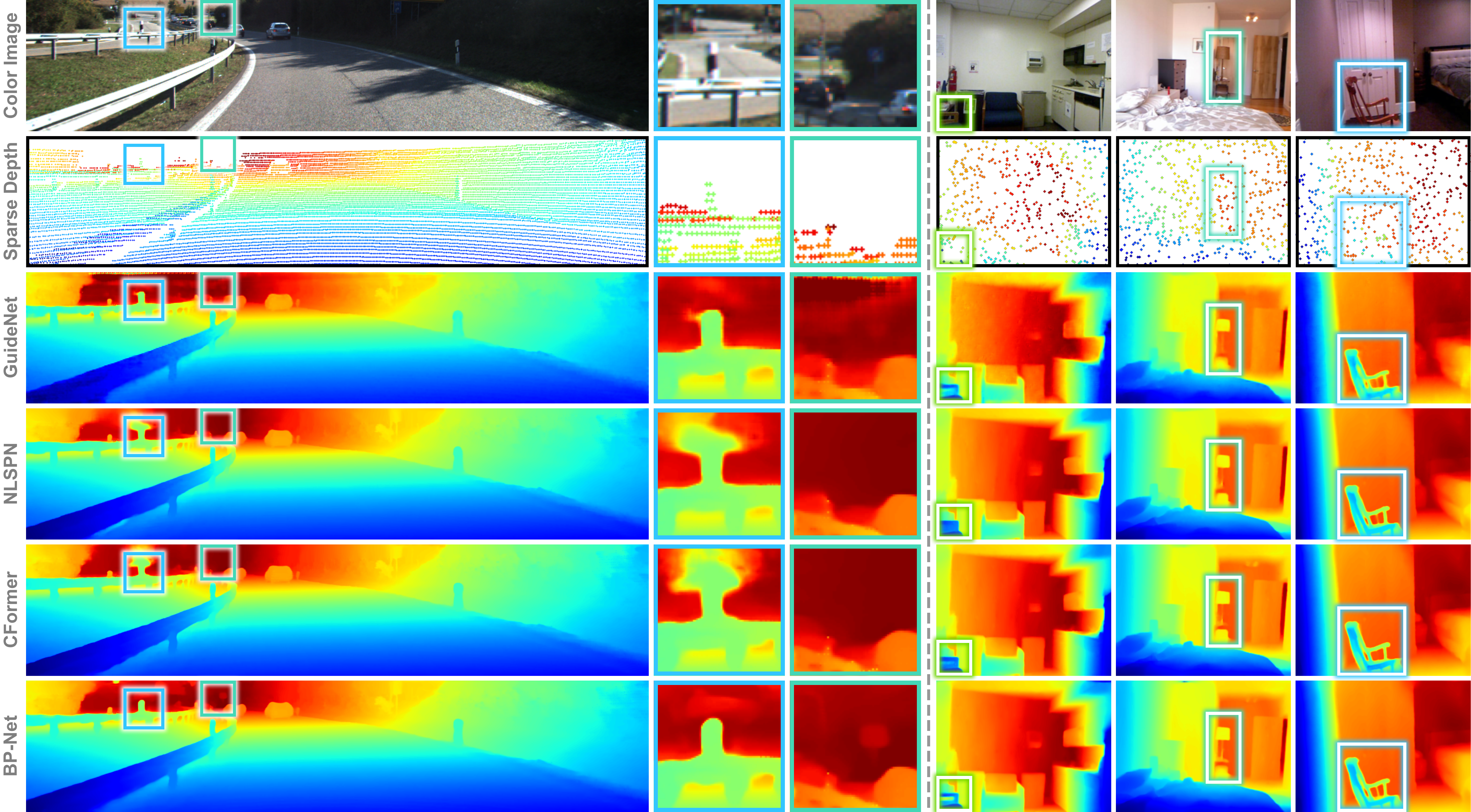}
    \vspace{-0.5em}
       \caption{Qualitative comparison with `GuideNet'~\cite{guidenet}, 
       `NLSPN'~\cite{nlspn} and `CFormer'~\cite{cformer} 
       on indoor and outdoor scenes.
       Sparse depth points are enlarged for better visualization.
       Our method is shown in the last row, with highlighted rectangles for easy comparison.}
       \label{fig:comparison}
       \vspace{-1em}
\end{figure*}

\cref{fig:comparison} provides visual comparisons with other SOTA open-source methods on the validation set of KITTI DC and the test set of NYUv2. We adopt the public code and best models from their authors to produce results and keep the sparse depth map the same for comparing methods. Our results listed in the last row exhibit clearer object boundaries and richer details, while other methods struggle to estimate accurate depth in these challenging regions.

\begin{figure}
    \begin{center}
    \includegraphics[width=0.98\columnwidth]{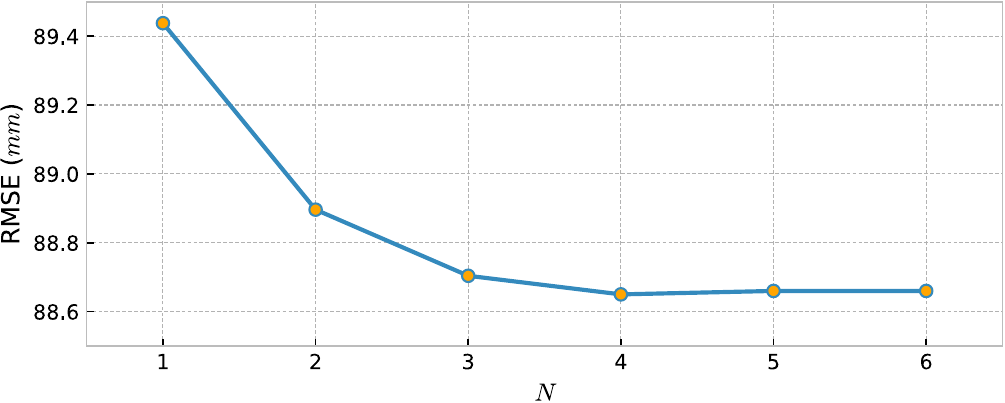}
    \end{center}
    \vspace{-2 em}
    \caption{The RMSE performance under different numbers of neighboring pixels.
    }\label{fig:validnum}
    \vspace{-1em}
\end{figure}

\subsection{Ablation Studies}
We conduct ablation studies on the NYUv2 dataset to reveal 
the effects of the different components in our method.

\subsubsection{Different Number for Neighboring Pixels} \label{sec:local_num}
In our bilateral propagation module, the target depth is propagated from $N$ nearest neighboring valid depth measurements according to our depth parametrization in~\cref{eq:dep_para}. We experiment with different values of $N$ and plot the corresponding RMSEs in \cref{fig:validnum}. 
While different choices of $N$ might slightly change the results at the \emph{Pre.} stage, they can all improve the \emph{MF.} stage for more effective depth completion.
Thus, our method is not very sensitive to this hyperparameter.
The performance gain is most noticeable when $N$ increases from 1 to 2, after which it gradually decreases and trends toward saturation. 
Finally, we choose $N=4$ in our method.

\begin{table}
    \begin{center}
        \scalebox{0.80}{
    \begin{tabular}{cccc|c}
        \hline
        \begin{tabular}[c]{@{}c@{}} Content \\ Propagation \end{tabular} & \begin{tabular}[c]{@{}c@{}} Spatial \\ Propagation \end{tabular} &  \begin{tabular}[c]{@{}c@{}}Multi-scale \\ Architecture \end{tabular} & \begin{tabular}[c]{@{}c@{}}Multi-scale \\ Loss \end{tabular} & \begin{tabular}[c]{@{}c@{}}RMSE$\downarrow$ \\ ($mm$)\end{tabular}  \\ 
        \hline
        \checkmark &  &  \checkmark    & \checkmark & 90.40   \\
         & \checkmark & \checkmark   & \checkmark & 90.93    \\
        \checkmark & \checkmark &    &  & 91.94   \\
        \checkmark & \checkmark & \checkmark   &  & 92.46    \\ \hline
        \checkmark & \checkmark & \checkmark   & \checkmark & 88.69   \\
        \hline
    \end{tabular}
        }
        \vspace{-0.5em}
        \caption{Ablation studies on NYUv2 dataset.}~\label{tab:ablation}
    \end{center}
    \vspace{-2em}
\end{table}

\subsubsection{Effect of Bilateral Parameters Generation}
Our bilateral propagation module is a non-linear model, whose coefficients are dynamically generated depending on image content and spatial distance as defined in~\cref{eq:para_gen}. To verify the contribution of these terms, We train a variant named \emph{content propagation}, where the coefficients depend only on $I_{i}, I_{j}, S_{j}$. We further train another variant named \emph{spatial propagation}, where the coefficients depend only on $\mathcal{O}_{ij}$. \cref{tab:ablation} lists the performance of these two variants and reveals that both \emph{content propagation} and \emph{spatial propagation} are useful, and removing any of them decreases the performance.

\subsubsection{Effect of Multi-Scale Depth Estimatation}
We also verify the effectiveness of our multi-scale scheme, which estimates dense depth in a multi-scale architecture with a multi-scale loss. We train a variant by replacing the multi-scale loss with a loss only on the original scale and train another variant by further replacing the multi-scale architecture with depth estimation only on the original scale. 
Based on the results listed in~\cref{tab:ablation}, we can see the integration of both multi-scale architecture and multi-scale loss enhances system performance 
while merely introducing the multi-scale architecture without incorporating multi-scale loss decreases the performance.
In BP-Net, multi-scale architecture facilitates coarse-to-fine depth estimation, where 
low-resolution results are upsampled to guide the propagation in high-resolution.
Consequently, multi-scale loss provides direct supervision at each resolution level, making the low-resolution results meaningful and improving the guidance effectively.
Omitting the multi-scale loss might make the network ignore low-resolution results, and increase training difficulty.

\subsubsection{Effect of 3-stage Depth Estimation}

At each scale, dense depth is estimated via three sequential stages, consisting of pre-processing (\emph{Pre.}), multi-modal fusion (\emph{MF.}), and post-processing (\emph{Post.}). We train different variants with different combinations of these three stages and analyze the performance gain on RMSE and extra cost on parameters (\emph{Params.}), Multiply-adds operations (\emph{Madds.}), GPU RAM and runtime. As shown in \cref{tab:3stages}, combining \emph{MF.} with \emph{Pre.} can achieve $2.12 mm$ performance gain on RMSE at the cost of $1.11 M$ more \emph{Params.} and $7.39 G$ more \emph{Madds.}
As a comparison, combining \emph{MF.} with \emph{Post.} produces less performance gain with more cost on \emph{Params.} and \emph{Madds.} In addition, \emph{Pre.+MF.} achieves better RMSE than \emph{MF.+ Post.}  with less GPU RAM and runtime. And combining all three stages achieves the best RMSE.
These comparison verify the effectiveness of our 3-stage design and prove the importance of depth propagation at the preprocessing stage in contrast to the refinement stage.

\begin{table}
   \begin{center}
       \scalebox{0.53}{
   \begin{tabular}{ccc|ccccc}
       \hline
       Pre. & MF. & Post.  & \begin{tabular}[c]{@{}c@{}}RMSE$\downarrow$ \\ ($mm$)\end{tabular}   & \begin{tabular}[c]{@{}c@{}}Params.$\downarrow$ \\ ($M$)\end{tabular}    & \begin{tabular}[c]{@{}c@{}}Madds.$\downarrow$ \\ ($G$)\end{tabular} & \begin{tabular}[c]{@{}c@{}}GPU RAM.$\downarrow$ \\ ($MB$)\end{tabular} & \begin{tabular}[c]{@{}c@{}}Runtime.$\downarrow$ \\ ($ms$)\end{tabular} \\ 
       \hline
       & \checkmark &      & $91.68(-0.00)$ & 76.15(+\, 0.00) & 105.76(+\, 0.00)  & 5206(+\, \,  0)  & \, 7.78(+\, 0.00) \\
       \checkmark & \checkmark &      & $89.56(-2.12)$ &77.26(+\, 1.11) & 113.15(+\, 7.39) &  5215(+\, \, 9)  & 12.95(+\, 5.17)  \\
        & \checkmark &  \checkmark  & $90.04(-1.68)$ &88.77(+12.62) & 129.73(+23.87) &  5302(+\, 96)  & 14.36(+\, 6.58)   \\
        \checkmark & \checkmark &  \checkmark  & $88.69(-2.99)$ &89.87(+13.72) & 137.12(+31.36) &  5312(+106)  & 19.68(+11.90)    \\

       \hline
   \end{tabular}
       }
       \vspace{-0.5em}
       \caption{Effect of 3-stage estimation.}\label{tab:3stages}
   \end{center}
   \vspace{-1em}
\end{table}

\subsection{Input Sparsity Analysis}

\begin{figure}
    \begin{center}
    \includegraphics[width=0.98\columnwidth]{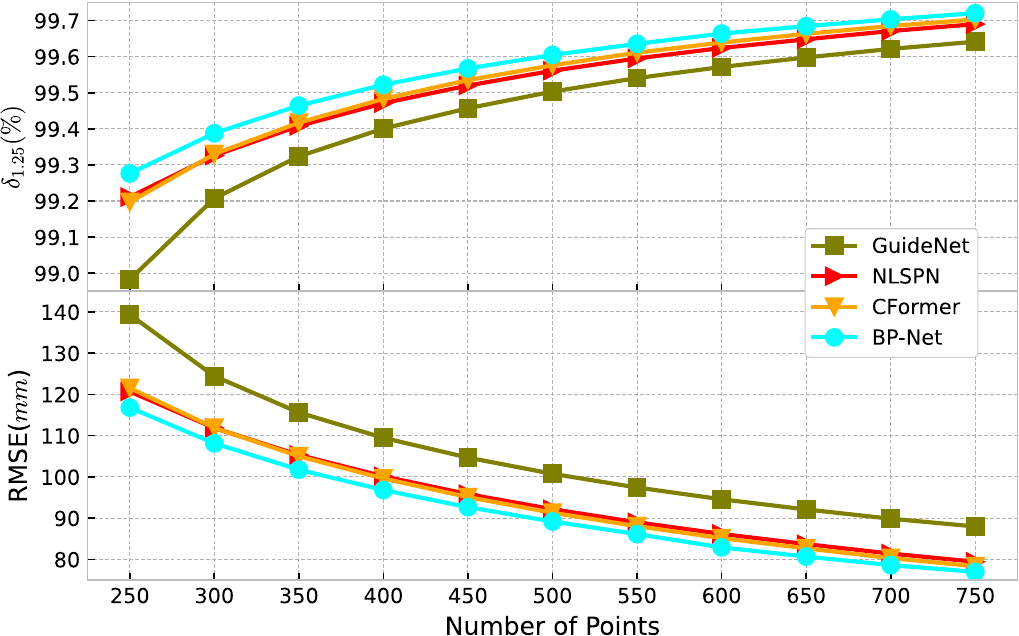}
    \end{center}
    \vspace{-1em}
    \caption{\textbf{The performance under various sparsity levels. }
    $\delta_{1.25}$ is plotted in the upper part, and RMSE is in the lower part.
    All comparing methods are directly evaluated without retraining.
    }\label{fig:sparsity}
    
\end{figure}

In real applications, the depth sparsity level might change, i.e. the number of points in the input sparse depth map might vary. We compare our method with others under various sparsity levels on the NYUv2 dataset. In this comparison, all models are trained with $500$ depth points and then evaluated under various sparsity levels ranging from $250$ to $750$ depth points.
For a thorough evaluation, given a sparsity level, each test image is sampled 100 times with different random seeds to generate the input sparse depth map. The performance of each method is averaged on these 100 randomly sampled inputs to reduce the potential bias due to random sampling. 
As shown in~\cref{fig:sparsity}, our method consistently achieves the lowest RMSE and the highest $\delta_{1.25}$  across various sparsity levels. The differences are more significant when the sparsity level is higher. This comparison demonstrates the strong generalization capability of our method across various sparsity levels.

\section{Conclusion}

This paper introduces a Bilateral Propagation Network (BP-Net) for depth completion.
BP-Net propagates depth at the earliest stage, rather than the refinement stage,
avoiding the following multi-modal stage from the sparsity problem.
The proposed bilateral propagation module can dynamically predict propagation coefficients conditioned on both \emph{radiometric difference} and \emph{spatial distance},
to enable depth propagation with the preference of nearest values on both domain and range.
Experimental results demonstrate the outstanding performance of BP-Net
and also suggest the importance of propagation at the earliest stage in contrast to the refinement stage.
BP-Net, mostly consisting of local propagation and convolution operations, 
may be limited to local structure and have difficulty with long-range information delivery.
Combining BP-Net with global and non-local operations, 
such as Transformer for multi-modal fusion~\cite{cformer} 
and non-local propagation for depth refinement~\cite{nlspn}, 
are directions for future works.

{
\noindent\textbf{Acknowledgements:}
This work was supported in part by National Natural Science Foundation of China (NSFC) under Grants 61973311, 62273353, and 62103431.
}

{
    \small
    \bibliographystyle{ieeenat_fullname}
    \bibliography{ref}
}

\end{document}



\clearpage
\setcounter{page}{1}
\maketitlesupplementary

\setcounter{section}{0}

\renewcommand\thesection{\Alph{section}}

This supplementary material provides additional information to complement the main paper. 
It includes implementation details of the proposed method in \cref{sec:app_method},
network architecture details in \cref{sec:net_arch},
 additional training details in \cref{sec:app_imp},
 descriptions of the adopted evaluation metrics in \cref{sec:app_eva}, 
 and further experimental results in Section~\ref{sec:app_results}.

\section{Additional Method Details}\label{sec:app_method}

\subsection{Inverse Projection Implementation}

Our method utilizes depth features in both prior encoding (Sec.~3.2.3) and multi-modal fusion (Sec.~3.3) of the main paper. 
These depth features are obtained by inverse projecting the depth map into camera space. 
This inverse projection technique has been proven beneficial for extracting 3D cues in a previous study~\cite{penet}. 
Specifically, the depth feature map $D$ is transformed from a single-channel depth map to a three-channel feature map, 
with each pixel coordinate $(x,y)$ represented by $(X,Y,Z)$, and can be written as:
\begin{equation}
    \begin{split}
     \begin{matrix}
        X_{x,y} = \frac{x-c_{x}}{f_{x}} D_{x,y},  \\
        Y_{x,y} = \frac{y-c_{y}}{f_{y}} D_{x,y},  \\
        Z_{x,y} = D_{x,y}.  \\
    \end{matrix}
    \end{split}
\end{equation}
Here,  $c_{x} , c_{y}, f_{x} , f_{y}$ are intrinsic parameters of a camera.
Our method employs a coarse-to-fine manner for depth estimation, 
where depth maps are generated at multiple scales. 
Thus we correspondingly adjust the intrinsic parameters used for depth feature generation at different scales.
Specially, for a scale $s$.
\begin{equation}
    \begin{split}
        c_{x}^{s} =\frac{c_{x}}{2^{s}},  
        c_{y}^{s} =\frac{c_{y}}{2^{s}},  \\
        f_{x}^{s} =\frac{f_{x}}{2^{s}},  
        f_{y}^{s} =\frac{f_{y}}{2^{s}}.  \\
    \end{split}
\end{equation}

\subsection{Image Encoding Implementation}

The proposed bilateral propagation module is arranged in a multi-scale scheme, 
with prior encodings from the corresponding resolution.
The image encoding $\mathcal{I}$ in scale $s$ can be written as:
\begin{equation}
    \begin{split}
    \mathcal{I}^{s}=\left\{ \begin{matrix}
        \mathbb{C}([\mathbf{I}^{s}, \mathbb{D}([\mathbf{F}^{s+1},\mathbf{D}^{s+1}])]),\, 0\le s<5,  \\
        \qquad \qquad  \; \ \mathbf{I}^{s},\qquad \qquad \qquad s=5.  \\
    \end{matrix} \right.
    \end{split}
\end{equation}
Here, for the lowest resolution with $s=5$,
we directly adopt the image feature $\mathbf{I}^{s}$ as image encoding.
Otherwise, to make image encoding representative, 
we concatenate the multi-modal fused feature $\mathbf{F}^{s+1}$ with depth feature $\mathbf{D}^{s+1}$ in scale $s+1$.
The depth feature $\mathbf{D}^{s+1}$ is achieved by inverse projecting estimated depth $D^{s+1}$ to camera space.
Then we utilize deconvolution operation $\mathbb{D}$ to upsample the concatenated feature map to scale $s$.
Finally, we concatenate the upsampled feature with $\mathbf{I}^{s}$, 
and adopt an extra convolution operation $\mathbb{C}$ to produce the image encoding $\mathcal{I}^{s}$.

\subsection{Weighted Pooling Implementation}
As explained in Sec.~3.2.3 of the main paper, 
we employ weighted pooling to downsample the sparse depth map. 
For a pixel $i$ at under scale $s$, the downsampled sparse depth map can be represented as:
\begin{equation}\label{eq:downsample}
    S_{i}^{s} = \frac{\sum\limits_{j=\mathcal{N}_{s}(i)}  w_{j}^{s} S_{j}}{\sum\limits_{j=\mathcal{N}_{s}(i)} w_{j}^{s} \mathbb{I}(S_{j})  +\epsilon},
\end{equation}
The weight map $w$ is estimated from image content and generated using an exponential layer to ensure positivity.
Thus, \cref{eq:downsample} can be explicitly formalized as
\begin{equation}\label{eq:downsample1}
    S_{i}^{s} = \frac{\sum\limits_{j=\mathcal{N}_{s}(i)}  e^{\hat{w}_{j}^{s}} S_{j}}{\sum\limits_{j=\mathcal{N}_{s}(i)} e^{\hat{w}_{j}^{s}} \mathbb{I}(S_{j})  +\epsilon},
\end{equation}
where, $\hat{w}$ is the generated weight map before exponential transform.
Directly implementing ~\cref{eq:downsample1} may have numerical risk on weights generation and gradients calculation.
In practice, we adopt an equivalent transformation that
\begin{equation}\label{eq:downsample2}
    \begin{split}
    S_{i}^{s} = \frac{\sum\limits_{j=\mathcal{N}_{s}(i)}  e^{\check{w}_{j}^{s}} S_{j}}{\sum\limits_{j=\mathcal{N}_{s}(i)} e^{\check{w}_{j}^{s}} \mathbb{I}(S_{j})  +\epsilon}, \\
    \check{w}_{j}^{s} = \hat{w}_{j}^{s} - \max\limits_{j=\mathcal{N}_{s}(i)}\hat{w}_{j}^{s}.
    \end{split}
\end{equation}
Here, by reducing the maximum value in $\mathcal{N}_{s}(i)$ for each pixel $j$,
$\check{w}_{j}^{s}$ is less equal than $0$, avoiding the potential numerical stability issue in implementation.

\section{Network Architecture}\label{sec:net_arch}

The overview of our BP-Net is depicted in Fig.~2 of the main paper.
We show the detailed architecture in \cref{tab:ablation} with images of $320 \times 256$ as input.
Here, symbols are consistent with the main paper, \eg ${{D}'}^{5}$ denotes the propagated depth map from bilateral propagation module in scale $5$.
Note that only main operators are listed in this table, and some trivial operations, \eg converting $S^{4}$ to $\mathcal{S}^{4}$ by inverse projection, are omitted for clarity.

\begin{table}
    \begin{center}
        \scalebox{0.80}{
    \begin{tabular}{c|c|c|c}
        \hline
        Output & Input & Operator  &  Output Size\\ 
        \hline
        $\mathbf{I}^{0}$ &  $I$    & Basic2D + ResBlock $\times 2 $  & $(\ \, 32, 256, 320)$ \\ \hline
        $\mathbf{I}^{1}$ &  $\mathbf{I}^{0}$    & ResBlock $\times 2 $  & $( \ \, 64, 128, 160)$ \\ \hline
        $\mathbf{I}^{2}$ &  $\mathbf{I}^{1}$    & ResBlock $\times 2 $  & $( 128, \ \, 64, \ \, 80)$ \\ \hline
        $\mathbf{I}^{3}$ &  $\mathbf{I}^{2}$   & ResBlock $\times 2 $  & $(256, \ \, 32, \ \, 40)$ \\ \hline
        $\mathbf{I}^{4}$ &  $\mathbf{I}^{3}$   & ResBlock $\times 2 $  & $(256, \ \, 16, \ \, 20)$ \\ \hline
        $\mathbf{I}^{5}$ &  $\mathbf{I}^{4}$  & ResBlock $\times 2 $  & $(256, \ \, \ \, 8, \ \, 10)$ \\ \hline
        $S^{0}$ &  $S$    & Identity  & $(\ \, \ \, 1, 256, 320)$ \\ \hline
        $S^{1}$ &  $\mathbf{I}^{1}, S$    & Weighted Pooling  & $(\ \, \ \, 1, 128, 160)$ \\ \hline
        $S^{2}$ &  $\mathbf{I}^{2}, S$    & Weighted Pooling  & $(\ \, \ \, 1, \ \, 64, \ \, 80)$ \\ \hline
        $S^{3}$ &  $\mathbf{I}^{3}, S$    & Weighted Pooling  & $(\ \, \ \, 1, \ \, 32, \ \, 40)$ \\ \hline
        $S^{4}$ &  $\mathbf{I}^{4}, S$    & Weighted Pooling  & $(\ \, \ \, 1, \ \, 16, \ \, 20)$ \\ \hline
        $S^{5}$ &  $\mathbf{I}^{5}, S$    & Weighted Pooling  & $(\ \, \ \, 1,\ \, \ \, 8, \ \, 10)$ \\ \hline
        $\mathcal{I}^{5}$ & $\mathbf{I}^{5}$    & Identity  & $(256, \ \, \ \, 8, \ \, 10)$  \\ \hline
        ${{D}'}^{5}$ &  $\mathcal{I}^{5}, \mathcal{S}^{5}, \mathcal{O}^{5} $    & \begin{tabular}[c]{@{}c@{}} \emph{Pre.} \\ \hline (Linear + BN + GeLU) $ \times 4 $ \end{tabular}  & $(\ \, \ \, 1,\ \, \ \, 8, \ \, 10)$ \\ \hline
        $\mathbf{F}^{5}$ &  $\mathcal{I}^{5}, \mathcal{S}^{5} $  & \begin{tabular}[c]{@{}c@{}} \emph{MF.} \\ \hline Basic2D + ResBlock $\times 2 \times 1 $ \end{tabular}  & $(256,\ \, \ \, 8, \ \, 10)$ \\ \hline
        ${{D}''}^{5}$ &  $\mathbf{F}^{5}, {{D}'}^{5} $    & Conv + Add  & $(\ \, \ \, 1,\ \, \ \, 8, \ \, 10)$ \\ \hline
        ${D}^{5}$ &  $S^{5},{{D}''}^{5} $    &  \begin{tabular}[c]{@{}c@{}} \emph{Post.} \\ \hline Conv $\times 3 \times 2 $ \end{tabular}    & $(\ \, \ \, 1,\ \, \ \, 8, \ \, 10)$ \\ \hline
        $\mathcal{I}^{4}$ & $\mathbf{I}^{4}, \mathbf{F}^{5}, {D}^{5}$    & Deconv + Conv  & $(256, \ \, 16, \ \, 20)$   \\ \hline
        ${{D}'}^{4}$ &  $\mathcal{I}^{4}, \mathcal{S}^{4}, \mathcal{O}^{4}$    & \begin{tabular}[c]{@{}c@{}} \emph{Pre.} \\ \hline (Linear + BN + GeLU) $ \times 4 $ \end{tabular}  & $(\ \, \ \, 1, \ \, 16, \ \, 20)$ \\ \hline
        $\mathbf{F}^{4}$ &  $\mathcal{I}^{4}, \mathcal{S}^{4} $  & \begin{tabular}[c]{@{}c@{}} \emph{MF.} \\ \hline Basic2D + ResBlock $\times 2 \times 2 $ \end{tabular}  & $(256, \ \, 16, \ \, 20)$ \\ \hline
        ${{D}''}^{4}$ &  $\mathbf{F}^{4},{{D}'}^{4} $    & Conv + Add  & $(\ \, \ \, 1, \ \, 16, \ \, 20)$ \\ \hline
        ${D}^{4}$ &  $S^{4},{{D}''}^{4} $    &  \begin{tabular}[c]{@{}c@{}} \emph{Post.} \\ \hline Conv $\times 3 \times 4 $ \end{tabular}    & $(\ \, \ \, 1, \ \, 16, \ \, 20)$ \\ \hline
        $\mathcal{I}^{3}$ & $\mathbf{I}^{3}, \mathbf{F}^{4}, {D}^{4}$    & Deconv + Conv  & $(256, \ \, 32, \ \, 40)$  \\ \hline
        ${{D}'}^{3}$ &  $\mathcal{I}^{3}, \mathcal{S}^{3}, \mathcal{O}^{3} $    & \begin{tabular}[c]{@{}c@{}} \emph{Pre.} \\ \hline (Linear + BN + GeLU) $ \times 4 $ \end{tabular}  & $(\ \, \ \, 1, \ \, 32, \ \, 40)$ \\ \hline
        $\mathbf{F}^{3}$ &  $\mathcal{I}^{3}, \mathcal{S}^{3} $  & \begin{tabular}[c]{@{}c@{}} \emph{MF.} \\ \hline Basic2D + ResBlock $\times 2 \times 3 $ \end{tabular}  & $(256, \ \, 32, \ \, 40)$ \\ \hline
        ${{D}''}^{3}$ &  $\mathbf{F}^{3},{{D}'}^{3} $    & Conv + Add  & $(\ \, \ \, 1, \ \, 32, \ \, 40)$ \\ \hline
        ${D}^{3}$ &  $S^{3},{{D}''}^{3} $    &  \begin{tabular}[c]{@{}c@{}} \emph{Post.} \\ \hline Conv $\times 3 \times 6 $ \end{tabular}    & $(\ \, \ \, 1, \ \, 32, \ \, 40)$ \\ \hline
        $\mathcal{I}^{2}$ & $\mathbf{I}^{2}, \mathbf{F}^{3}, {D}^{3}$    & Deconv + Conv  & $( 128, \ \, 64, \ \, 80)$  \\ \hline
        ${{D}'}^{2}$ &  $\mathcal{I}^{2}, \mathcal{S}^{2}, \mathcal{O}^{2} $    & \begin{tabular}[c]{@{}c@{}} \emph{Pre.} \\ \hline (Linear + BN + GeLU) $ \times 4 $ \end{tabular}  & $(\ \, \ \, 1, \ \, 64, \ \, 80)$ \\ \hline
        $\mathbf{F}^{2}$ &  $\mathcal{I}^{2}, \mathcal{S}^{2} $  & \begin{tabular}[c]{@{}c@{}} \emph{MF.} \\ \hline Basic2D + ResBlock $\times 2 \times 4 $ \end{tabular}  & $( 128, \ \, 64, \ \, 80)$ \\ \hline
        ${{D}''}^{2}$ &  $\mathbf{F}^{2},{{D}'}^{2} $    & Conv + Add  & $(\ \, \ \, 1, \ \, 64, \ \, 80)$ \\ \hline
        ${D}^{2}$ &  $S^{2},{{D}''}^{2} $    &  \begin{tabular}[c]{@{}c@{}} \emph{Post.} \\ \hline Conv $\times 3 \times 8 $ \end{tabular}    & $(\ \, \ \, 1, \ \, 64, \ \, 80)$ \\ \hline
        $\mathcal{I}^{1}$ & $\mathbf{I}^{1}, \mathbf{F}^{2}, {D}^{2}$    & Deconv + Conv  & $( \ \, 64, 128, 160)$  \\ \hline
        ${{D}'}^{1}$ &  $\mathcal{I}^{1}, \mathcal{S}^{1}, \mathcal{O}^{1} $    & \begin{tabular}[c]{@{}c@{}} \emph{Pre.} \\ \hline (Linear + BN + GeLU) $ \times 4 $ \end{tabular}  & $(\ \, \ \, 1, 128, 160)$ \\ \hline
        $\mathbf{F}^{1}$ &  $\mathcal{I}^{1}, \mathcal{S}^{1} $  & \begin{tabular}[c]{@{}c@{}} \emph{MF.} \\ \hline Basic2D + ResBlock $\times 2 \times 5 $ \end{tabular}  & $( \ \, 64, 128, 160)$ \\ \hline
        ${{D}''}^{1}$ &  $\mathbf{F}^{1},{{D}'}^{1} $    & Conv + Add  & $(\ \, \ \, 1, 128, 160)$ \\ \hline
        ${D}^{1}$ &  $S^{1},{{D}''}^{1} $    &  \begin{tabular}[c]{@{}c@{}} \emph{Post.} \\ \hline Conv $\times 3 \times 10 $ \end{tabular}    & $(\ \, \ \, 1, 128, 160)$ \\ \hline
        $\mathcal{I}^{0}$ & $\mathbf{I}^{0}, \mathbf{F}^{1}, {D}^{1}$    & Deconv + Conv  & $(\ \, 32, 256, 320)$  \\ \hline
        ${{D}'}^{0}$ &  $\mathcal{I}^{0}, \mathcal{S}^{0}, \mathcal{O}^{0} $    & \begin{tabular}[c]{@{}c@{}} \emph{Pre.} \\ \hline (Linear + BN + GeLU) $ \times 4 $ \end{tabular}  & $(\ \, \ \, 1, 256, 320)$ \\ \hline
        $\mathbf{F}^{0}$ &  $\mathcal{I}^{0}, \mathcal{S}^{0} $  & \begin{tabular}[c]{@{}c@{}} \emph{MF.} \\ \hline Basic2D + ResBlock $\times 2 \times 6 $ \end{tabular}  & $(\ \, 32, 256, 320)$ \\ \hline
        ${{D}''}^{0}$ &  $\mathbf{F}^{0},{{D}'}^{0} $    & Conv + Add  & $(\ \, \ \, 1, 256, 320)$ \\ \hline
        ${D}^{0}$ &  $S^{0},{{D}''}^{0} $    &  \begin{tabular}[c]{@{}c@{}} \emph{Post.} \\ \hline Conv $\times 3 \times 12 $ \end{tabular}    & $(\ \, \ \, 1, 256, 320)$ \\ 
        \hline
    \end{tabular}
        }
        \caption{Detailed Architecture of BP-Net.}~\label{tab:ablation}
    \end{center}
\end{table}

\section{Additional Training Details}\label{sec:app_imp}

When training on KITTI dataset, we randomly crop image to $256\times 1216$ for training.
Following previous works~\cite{cformer,guidenet}, we adopt random horizontal flip, color jitter, 
and normalization as data augmentation.
When training on NYUv2 dataset,
we follow data augmentation in previous works~\cite{nlspn,cformer},
including random horizontal flip, random crop, random rotation, random resize, 
color jitter and normalization.
We apply data augmentation on color image and sparse depth map,
and adjust the camera intrinsic parameters correspondingly.

\section{Details on Evaluation Metrics}\label{sec:app_eva}

We verify our method on both indoor and outdoor scenes with standard evaluation metrics.
For indoor scene, root mean squared error (RMSE), mean absolute relative error (REL), and $\delta_{\theta}$  are chosen as evaluation metrics.
For outdoor scene, the standard evaluation metrics are root mean squared error (RMSE), mean absolute error (MAE), root mean squared error of the inverse depth (iRMSE), and mean absolute error of the inverse depth (iMAE).
These evaluation metrics are firstly calculated on each sample and then averaged among samples. 
And for each sample, they can be written as: 
\begin{equation}
    \begin{split}
        \begin{matrix}
            RMSE  = (\frac{1}{n}\sum_{i \in \mathcal{P}_{v}}(D_{i}^{gt}-D_{i})^2)^{\frac{1}{2}}, \\
            iRMSE  = (\frac{1}{n}\sum_{i \in \mathcal{P}_{v}}(\frac{1}{D_{i}^{gt}}-\frac{1}{D_{i}})^2)^{\frac{1}{2}}, \\
            MAE = \frac{1}{n}\sum_{i \in \mathcal{P}_{v}}\left | D_{i}^{gt}-D_{i} \right |, \\
            iMAE  = \frac{1}{n}\sum_{i \in \mathcal{P}_{v}}\left | \frac{1}{D_{i}^{gt}}-\frac{1}{D_{i}} \right |, \\
            REL = \frac{1}{n}\sum_{i \in \mathcal{P}_{v}}\frac{\left | D_{i}^{gt}-D_{i} \right |}{D_{i}^{gt}}, \\
            \delta_{\theta}  = \frac{1}{n}\sum_{i \in \mathcal{P}_{v}}\left |  \left \{ max ( \frac{D_{i}^{gt}}{D_{i}} ,\frac{D_{i}}{D_{i}^{gt}})<\theta \right \} \right |. \\
        \end{matrix}
    \end{split}
\end{equation}
Here, $\mathcal{P}_{v}$ is the set of pixels with valid ground truth, and $n=\left |  \mathcal{P}_{v} \right |$ is the size of the set.

\section{Additional Experimental Results}\label{sec:app_results}
\begin{figure*}
    \begin{center} 
       \includegraphics[width=1.0\textwidth]{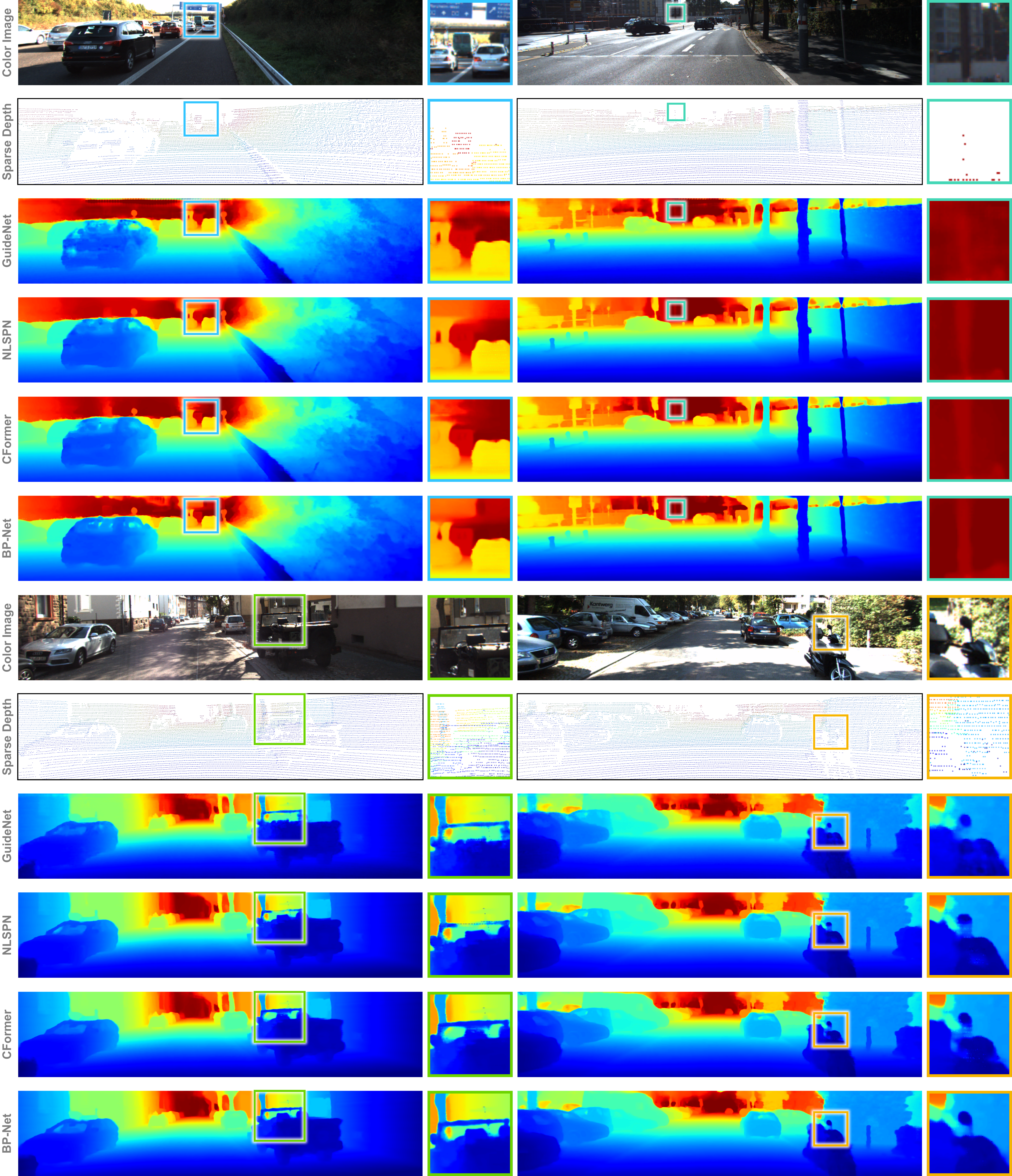}
    \end{center}
       \caption{\textbf{Additional qualitative results on KITTI dataset.}
       }\label{fig:more_comp_kitti}
\end{figure*}

\begin{figure*}
    \begin{center} 
       \includegraphics[width=1.0\textwidth]{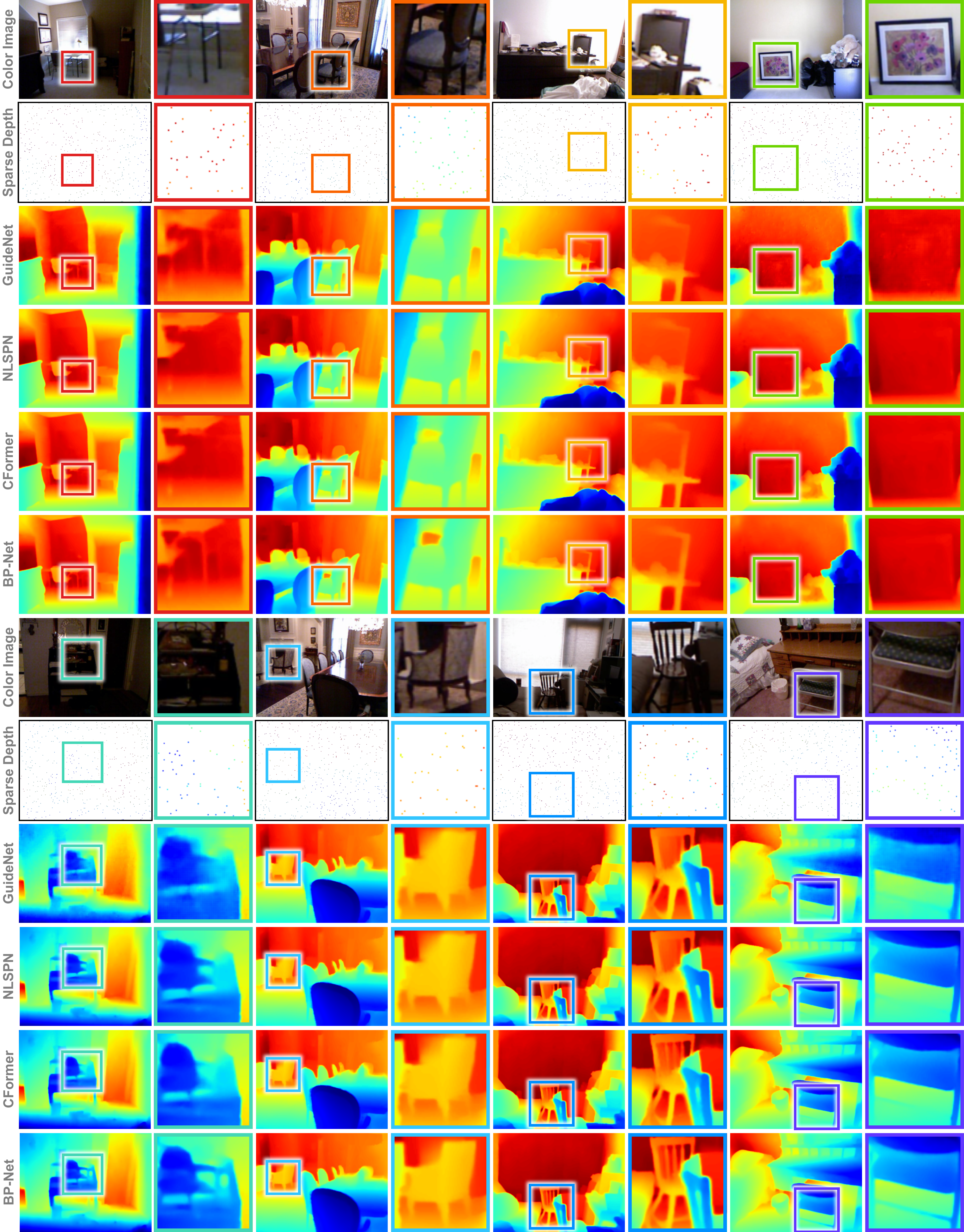}
    \end{center}
       \caption{\textbf{Additional qualitative results on NYUv2 dataset.}
       }\label{fig:more_comp_nyu}
\end{figure*}

Due to space limitation, we only show limited comparison results in Tab.~1 and Fig.~5 of the main paper.
Here, we list more performance evaluation on outdoor scene and indoor scene 
in \cref{tab:comparison_kitti} and  \cref{tab:comparison_nyu} respectively.
We also show more qualitative results on outdoor scene and indoor scene 
in \cref{fig:more_comp_kitti} and  \cref{fig:more_comp_nyu} respectively.

\begin{table}[!ht]
    \begin{center}
    \scalebox{0.9}{
    \begin{tabular}{@{\extracolsep{3pt}}lcccc@{}} \toprule
     & \begin{tabular}[c]{@{}c@{}}\textbf{RMSE}$\downarrow$ \\ ($mm$)\end{tabular}  
     & \begin{tabular}[c]{@{}c@{}}MAE$\downarrow$ \\ ($mm$)\end{tabular}  
     & \begin{tabular}[c]{@{}c@{}}iRMSE$\downarrow$ \\ ($1/km$)\end{tabular} 
     & \begin{tabular}[c]{@{}c@{}}iMAE$\downarrow$ \\ ($1/km$)\end{tabular}  \\ \midrule
    S2D~\cite{s2d} & 814.73 & 249.95 & 2.80 & 1.21  \\
    CSPN~\cite{cspn} &  1019.64 & 279.46 & 2.93 & 1.15  \\
    DeepLiDAR~\cite{deeplidar} & 758.38 & 226.50 & 2.56 & 1.15 \\
    FuseNet~\cite{2d3d} &  752.88 & 221.19 & 2.34 & 1.14 \\
    CSPN++~\cite{cspn++}  & 743.69 & 209.28 & 2.07 & 0.90  \\
    GuideNet~\cite{guidenet} & 736.24 & 218.83 & 2.25 & 0.99 \\
    FCFR~\cite{penet} & 735.81&  217.15&  2.20&  0.98  \\
    ACMNet~\cite{acmnet} & 744.91 & 206.09 & 2.08 &  0.90 \\
    NLSPN~\cite{nlspn} & 741.68 & 199.59 & 1.99 & 0.84  \\
    PENet~\cite{penet} & 730.08 & 210.55 & 2.17 & 0.94 \\
    RigNet~\cite{rignet} & 712.66 & 203.25 & 2.08 & 0.90  \\
    DySPN~\cite{dyspn} & 709.12 & 192.71 & 1.88 & 0.82 \\
    BEV@DC~\cite{bevdc} & 697.44 & 189.44 & 1.83 & 0.82  \\
    CFormer~\cite{cformer} &  708.87 & 203.45 & 2.01 & 0.88  \\
    LRRU~\cite{lrru} & 696.51 & \textbf{189.96} & 1.87 & \textbf{0.81} \\
    \midrule
    BP-Net & \textbf{684.90} & 194.69 & \textbf{1.82} & 0.84  \\ \bottomrule
    \end{tabular}
    }
    \caption{{\bf Performance on KITTI dataset.}
    Results are evaluated by the KITTI testing server and ranked by the RMSE (in $mm$).
    The best result under each criterion is in \textbf{bold}.
    } \label{tab:comparison_kitti}
    \end{center}
 \end{table}

\begin{table}[!ht]
    \begin{center}
    \scalebox{0.80}{
    \begin{tabular}{@{\extracolsep{3pt}}lccccc@{}} \toprule
     & \begin{tabular}[c]{@{}c@{}} \textbf{RMSE}$\downarrow$ \\ ($m$)\end{tabular}  
     & \begin{tabular}[c]{@{}c@{}}REL$\downarrow$ \\ $\ $ \end{tabular}
     & \begin{tabular}[c]{@{}c@{}}$\delta_{1.25}\uparrow$ \\ ($\%$)\end{tabular}
     & \begin{tabular}[c]{@{}c@{}}$\delta_{1.25^{2}}\uparrow$ \\ ($\%$)\end{tabular}
     & \begin{tabular}[c]{@{}c@{}}$\delta_{1.25^{3}}\uparrow$ \\ ($\%$)\end{tabular}
      \\ \midrule
    S2D~\cite{s2d} &  0.230 & 0.044 &  97.1 &99.4 & 99.8 \\
    CSPN~\cite{cspn} &   0.117 & 0.016 & 99.2& \textbf{99.9}& \textbf{100.0}\\
    DeepLiDAR~\cite{deeplidar} &   0.115 & 0.022 & 99.3 & \textbf{99.9}& \textbf{100.0} \\
    CSPN++~\cite{cspn++}  & 0.115 & -- & -- & --& -- \\
    DepthNormal~\cite{normal_constraint}  & 0.112 & 0.018 & 99.5 & \textbf{99.9}& \textbf{100.0} \\
    GuideNet~\cite{guidenet} &  0.101 &0.015&99.5& \textbf{99.9}& \textbf{100.0}\\
    FCFR~\cite{penet} &  0.106 & 0.015& 99.5& \textbf{99.9}& \textbf{100.0} \\
    ACMNet~\cite{acmnet} &  0.105 & 0.015 & 99.4 & \textbf{99.9}& \textbf{100.0}\\
    TWISE~\cite{TWISE} &  0.097 & 0.013 & \textbf{99.6} & \textbf{99.9}& \textbf{100.0}\\
    NLSPN~\cite{nlspn} &   0.092 & 0.012 & \textbf{99.6}& \textbf{99.9}& \textbf{100.0} \\
    RigNet~\cite{rignet} &  0.090 & 0.013 & \textbf{99.6} & \textbf{99.9}& \textbf{100.0} \\
    DySPN~\cite{dyspn} &  0.090 &  0.012 & \textbf{99.6}& \textbf{99.9}& \textbf{100.0} \\
    GraphCSPN~\cite{graphcspn} & 0.090 &  0.012 & \textbf{99.6}& \textbf{99.9}& \textbf{100.0} \\
    BEV@DC~\cite{bevdc} &  \textbf{0.089} &0.012& \textbf{99.6}& \textbf{99.9}& \textbf{100.0} \\
    CFormer~\cite{cformer} &   0.090 &0.012&-- & --& --\\
    LRRU~\cite{lrru} &   0.091 & \textbf{0.011} & \textbf{99.6}& \textbf{99.9}& \textbf{100.0} \\
    \midrule
    BP-Net &  \textbf{0.089} & 0.012 & \textbf{99.6}& \textbf{99.9}& \textbf{100.0} \\ \bottomrule
    \end{tabular}
    }
    \caption{{\bf Performance on NYUv2 datasets.}
    For the NYUv2 dataset, authors report their performance on the official test set in their papers.
    The best result under each criterion is in \textbf{bold}.
    } \label{tab:comparison_nyu}
    \end{center}
\end{table}

{
    \small
    \bibliographystyle{ieeenat_fullname}
    \bibliography{ref}
}